\documentclass[letterpaper, 10 pt, conference]{ieeeconf}  

\IEEEoverridecommandlockouts                              

\title{\LARGE \bf
AZTR: Aerial Video Action Recognition with Auto Zoom and Temporal Reasoning
}

\author{Xijun Wang*$^{1}$, Ruiqi Xian*$^{2}$ , Tianrui Guan$^{1}$, Celso M. de Melo$^{3}$, Stephen M. Nogar$^{3}$,\\ Aniket Bera$^{4}$ and Dinesh Manocha$^{1}$
\thanks{*These authors contributed equally}
\thanks{$^{1}$Authors are with Dept. of Computer Science, University of Maryland, College Park, MD, USA.
        {\tt\small xijun@umd.edu}}%
\thanks{$^{2}$Author is with the Dept. of Electrical and Computer Engineering, University of Maryland, College Park, MD, USA
        {\tt\small rxian@umd.edu}}%
\thanks{$^{3}$Author is with the Computational and Information Sciences Directorate, DEVCOM U.S. Army Research Laboratory, Adelphi, MD, USA
        }%
 \thanks{$^{4}$Author is with the Dept. of Computer Science, Purdue University, West Lafayette, IN, USA.
        {\tt\small ab@cs.purdue.edu}}%
}

\usepackage{graphicx}
\usepackage{amsmath}
\usepackage{amssymb}
\usepackage{booktabs}

\usepackage[font=small,belowskip=-1.5pt]{caption}

\begin{document}

\maketitle
\thispagestyle{empty}
\pagestyle{empty}


\begin{abstract}

We propose a novel approach for aerial video action recognition. Our method is designed for videos captured using UAVs and can run on edge or mobile devices. We present a learning-based approach that uses customized auto zoom to automatically identify the human target and scale it appropriately. This makes it easier to extract the key features and reduces the computational overhead. We also present an efficient temporal reasoning algorithm to capture the action information along the spatial and temporal domains within a controllable computational cost. Our approach has been implemented and evaluated both on the desktop with high-end GPUs and on the low power Robotics RB5 Platform for robots and drones. In practice, we achieve 6.1-7.4\% improvement over SOTA in Top-1 accuracy on the RoCoG-v2 dataset, 8.3-10.4\% improvement on the UAV-Human dataset and 3.2\% improvement on the Drone Action dataset.

\end{abstract}

\section{Introduction}
\label{sec: introduction}
There is considerable interest in capturing aerial videos of humans using drones and UAVs (unmanned aerial vehicles). This gives rise to challenging problems related to detection, tracking, recognition, person re-detection and action recognition on aerial data~\cite{nguyen2022state}. In this paper, we mainly deal with the problem of aerial video recognition and develop solutions that can also work well on low-power or edge hardware.

Recently, many deep learning methods have been proposed for video action recognition. Despite the great success of those methods on ground camera videos, a large drop in accuracy is observed when directly applying them to videos captured using UAV cameras. This is due to the domain shift caused by different viewing angles and camera characteristics.
Some of the challenges arise due to:

 \textit{Small resolution:} The target human actors  appear significantly smaller in aerial data due to high camera altitude. A wider area of the background is covered from the air, occupying most of the pixels in the video frame, and a small fraction of pixels correspond to a human action.
 
 \textit{Multi-scale:} Depending on the flying altitude of the UAV, the human actor may appear dramatically different in terms of size and scale. Such discrepancy makes it hard to extract features for model training, decreasing the overall accuracy.
 
 \textit{Moving camera:} The location of the human actor in the video may continuously change due to the movement of the UAV. The motion of the UAV causes more background changes than human behavior variations. This makes the model infer more from the background than the human actor, especially in high-resolution videos.

Furthermore, generic action recognition methods are mostly designed for desktop or cloud GPUs, which have high memory or power requirements. They can not be deployed on mobile or edge devices or UAV platforms with reduced memory and a lack of support for complex arithmetic operations.
Current recognition methods for mobile platforms~\cite{pan2022edgevits}\cite{kondratyuk2021movinets}\cite{demir2021tinyvirat} are designed for ground camera videos and cannot achieve great performance on aerial data. Therefore, we need better techniques for aerial video action recognition.

\begin{figure*}[t]
    \centering
        \includegraphics[width=\linewidth]{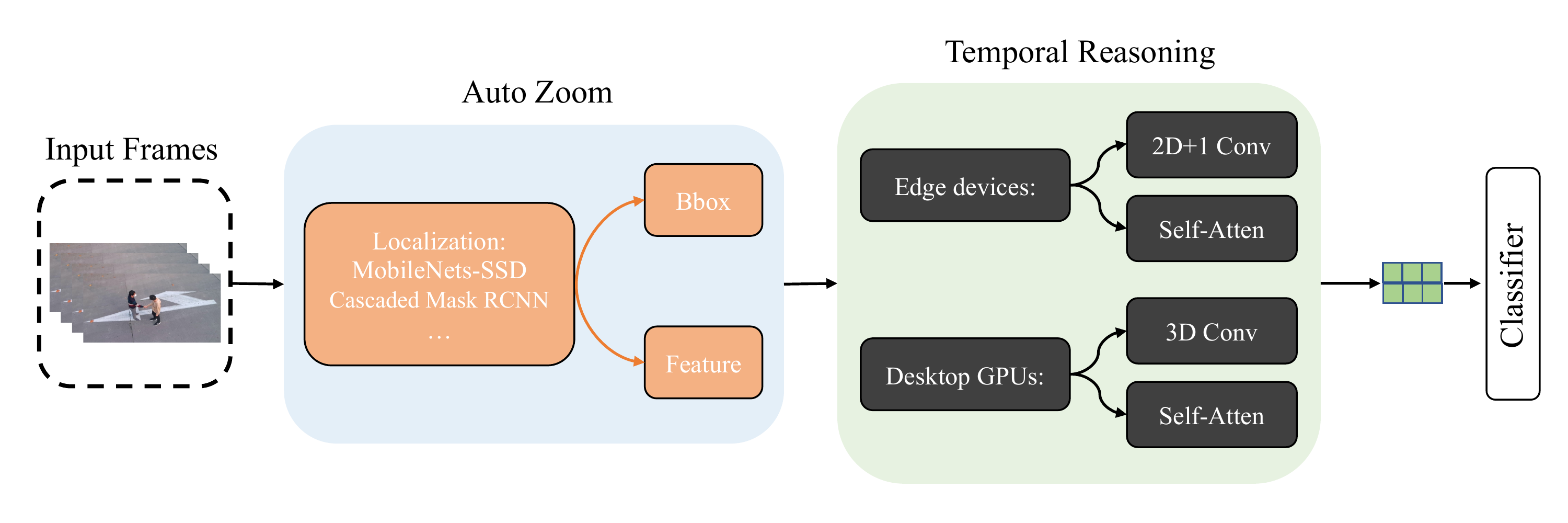}
        \caption{Our learning pipeline consists of the auto zoom learning algorithm and the temporal reasoning algorithm. For auto zoom learning, we offer different bounding box(bbox) and feature operations. Refer to Section ~\ref{sec: method} for details. For the temporal reasoning algorithm, we perform (2D+1) conv on edge devices,  3D conv on desktop GPUs, and self-attention (Atten) mechanism on both edge devices and desktop GPUs. Attention layers on desktop GPUs are deeper and wider. }
        \vspace{-15pt}
    \label{fig: intro}
\end{figure*}

\subsection{Main Contributions}
We present a novel deep learning method for video action recognition. This includes a new auto zoom algorithm that efficiently identifies the target and scales the target region to a size that fits in the memory of a given device or processor. Our auto zoom algorithm uses an efficient technique for small resolution, multiple scales and moving cameras. Additionally, from the long-range space-time relation aspect, we introduce a temporal reasoning algorithm to capture the action information. The novel components of our work include:
\begin{enumerate}

\item We design an auto zoom algorithm for aerial video action recognition models. It can efficiently apply autofocus, cropping and scaling strategy to obtain the key action information from the human actor. The focused view changes every frame according to the position of the human actor, which compromises the UAV's motion so that the human actor always appears in the center of the video. This brings less background noise and extracts more useful features for the human behavior analysis, making the model more robust. 

\item We present a temporal reasoning algorithm, which uses a combination of convolution and attention mechanism to achieve better accuracy. We perform 3D convolutions on high-end desktop GPUs and (2D+1) convolutions on low-power edge devices or UAV platforms to balance between accuracy and inference speed. The attention algorithm is composed of cross-attention and self-attention that provides both spatial and temporal representations, which has linear computational complexity. Overall, our formulation can acquire long-range spatial-temporal relationships, obtains better understandings of the actions across the video.

\end{enumerate}
Our learning method can be adapted to high-end GPUs as well as low-power edge devices. We implement and evaluate our approach on the edge device Qualcomm Robotics RB5 Platform (Kryo 585 CPU and Adreno 650 GPU) and desktop GPUs (Nvidia RTX A5000 GPUs). We have evaluated the results and observe 6.1-7.4\% improvement over SOTA in Top-1 accuracy on the RoCoG-v2 dataset, 8.3-10.4\% improvement on the UAV-Human dataset, 3.2\% improvement on the Drone Action dataset. Our method can achieve 40.2\% accuracy with an inference time of 56.5 ms on the RB5 CPU, which outperforms the SOTA MoViNets in terms of accuracy and speed.

\section{Related works}
\label{sec: related_work}

\subsection{Learning-based Methods for Aerial Video Recognition}
The recent developments in deep learning methods have resulted in improved performance of action recognition on ground-camera video datasets\cite{rezazadegan2017action, soans2020sa, van2020multi, massardi2020parc, yao2022pa,shao2018hierarchical,lea2016learning}. However, their accuracy is not good for aerial videos\cite{nguyen2022state}. \cite{geraldes2019uav,mliki2020human,mishra2020drone,mou2020event} use ResNet and InceptionNet to perform single-frame classification and fuse all the outputs for recognition. 
\cite{barekatain2017okutama,perera2019drone,perera2020multiviewpoint} exploit two-stream CNNs to utilize attributes from appearance and motion. 3D CNNs are also widely used for aerial action recognition. \cite{choi2020unsupervised,demir2021tinyvirat,li2021uav,mou2020event,sultani2021human} adopt I3D\cite{carreira2017quo} to extract spatial-temporal features for more accurate recognition. \cite{peng2020fully} improves the Inception-ResNet model with 3D convolutions to make the model more suitable for aerial video processing. Other techniques use transformer-based solutions. \cite{kothandaraman2022fourier} combines Fourier transform with an attention mechanism for better feature extraction. Our method can be used as an complement to those methods, increasing the overall accuracy on aerial action recognition.

\begin{figure*}[t]
    \centering
        \includegraphics[width=\linewidth]{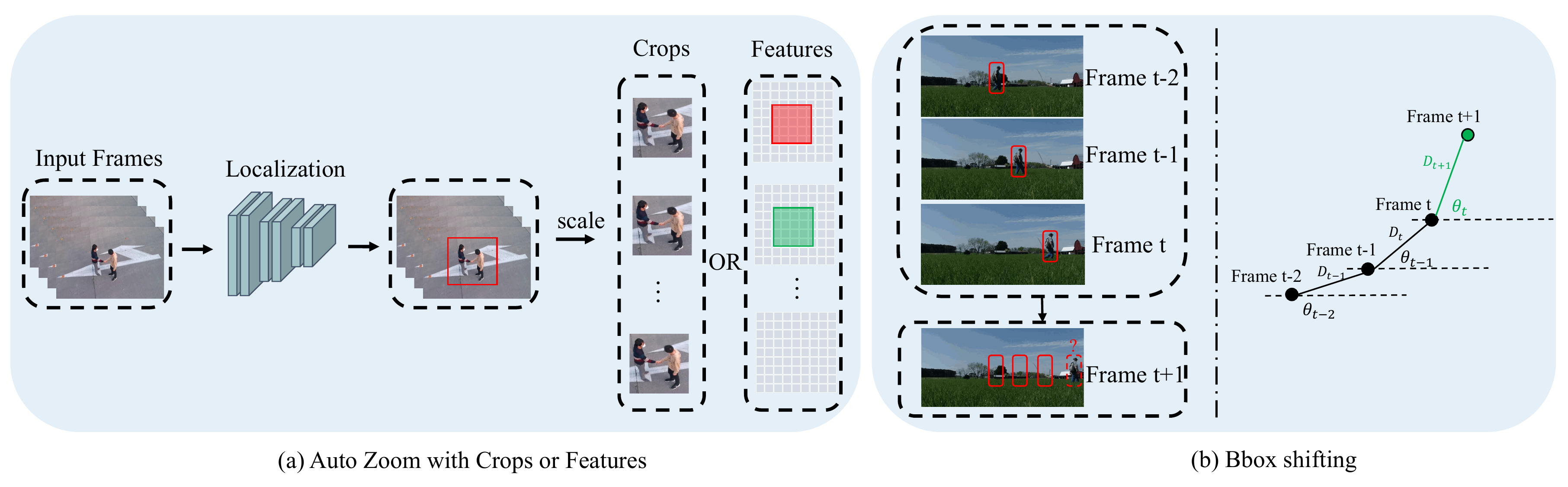}
        \caption{We designed two different auto zoom methods with crops or features, for high-end desktop and mobile or edge devices respectively. (a) For auto zoom with crops, we use a detector to get the target bounding box and crop it from the original frame, then scale the crop size. For the auto zoom with features, we use the features to generate the bounding boxes and classification. (b) We use the detector to generate bboxes on key frames to reduce the computational cost. We predict the bbox at the next key frame, and compare the location of predicted bbox and generated bbox to avoid incorrect detection results. Finally, we apply linear interpolation to generate the bbox between key frames. Details are shown in Section~\ref{sec: method}.}
    \label{fig: auto_zoom}
    \vspace{-10pt}
\end{figure*}

\subsection{UAV and Drone Datasets}
In order to stimulate further research, many UAV and drone datasets have been captured using affordable off-the-shelf drones. These include many public UAV datasets for most drone-based tasks like human detection, object tracking, human re-identification, and action recognition \cite{li2021uav,du2018unmanned,zhu2020detection,perera2018uav,perera2019drone,choi2020unsupervised,de2020vision}. Drone Action\cite{perera2019drone} is an outdoor drone video dataset providing 240 HD video clips recorded from low altitudes and at low speeds across 13 dynamic human actions. 
UAV-Human\cite{li2021uav} is a large benchmark for human behavior or action understanding with UAVs, which contains 67,248 multi-modal video sequences and 119 subjects for action recognition, 22,476 frames for pose estimation, 41,290 frames and 1,144 identities for person re-identification, and 22,263 frames for attribute recognition. 

\subsection{Activity Recognition on Edge Architectures}
Most deep learning methods use large and deep architectures to achieve higher accuracy, but this significantly increases the computational cost. As a result, these methods are used on desktop or cloud GPUs and are not practical for edge or low-power mobile devices. Improving the efficiency of video models with lightweight methods has gained increased attention~\cite{fan2019more,bhardwaj2019efficient,chen2018big,li2020smallbignet,tran2019video}. MobileNet\cite{howard2017mobilenets} and Yolo\cite{redmon2016you} are widely used networks for video modeling on edge devices. \cite{ding2020lightweight} proposes a lightweight action recognition model on UAVs using MobileNet with a focal loss and self-attention. \cite{kondratyuk2021movinets} presents Mobile Video Networks (MoViNets), a family of computational and memory-efficient video networks that can operate on streaming video for online inference. \cite{piergiovanni2022tiny} proposes Tiny Video Networks, which are automatically designed efficient video architectures. \cite{lin2019tsm} introduces the Temporal Shift Module(TSM), which shifts the channels along the temporal dimension, supporting both offline and online video recognition. Our auto zoom algorithm can be combined with any of these methods to achieve higher accuracy within limited resource requirement.

\section{Our Approach: AZTR}
\label{sec: method}
We propose a general aerial video action recognition learning method, that can automatically identify the human target, scale it
appropriately, and analyze the action by reasoning the input frames in the temporal dimension. To achieve those functions, we introduce 2 components: auto zoom algorithm and temporal reasoning algorithm that can be used by any video models to efficiently extract spatial and temporal features.

\subsection{Overall Learning Method}
As show in Fig ~\ref{fig: intro}, the input of our method are sampled video frames. Those frames will first use the auto zoom algorithm to get the key spatial information before using the temporal reasoning algorithm to get the action temporal information. 
 %
 For the auto zoom component, we use a target locator to obtain the sparse bounding box (bbox) or corresponding features. We can reduce the detection frequency by using bbox/feature shifting for sparse bboxes to save computational cost. Moreover, we use those bboxes to align the targets in the original video or those feature sequences for reasoning. For the temporal reasoning module, we use different methods for edge devices (with low memory and power requirements) and desktop GPUs. 2D+1 convolution and attention mechanism can be used in the temporal reasoning module on edge devices. Attention mechanism and 3D convolution are used to obtain high performance on desktop GPUs.
  
\begin{figure*}[h]
    \centering
        \includegraphics[width=\linewidth]{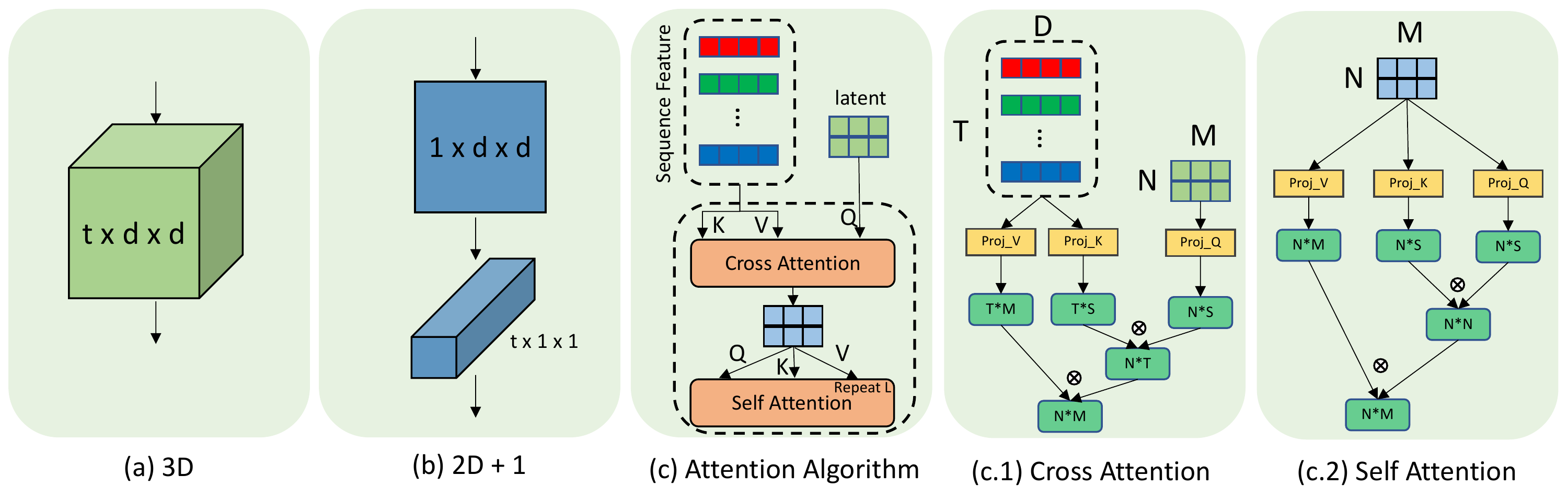}
        \caption{We use different combinations on desktop GPUs and edge devices between 2D+1 convolution, 3D convolution, and efficient transformer for temporal reasoning. The efficient transformer based algorithm has two components, the cross attention is used to map the input sequences to a new sequence with a specific size according the computational cost requirement. The self attention is the normal component from transformers.}
    \label{fig: temporal_reasoning}
\end{figure*}

\subsection{Auto Zoom}
We introduce an auto zoom algorithm for aerial video action recognition models. As shown in Fig.\ref{fig: auto_zoom}, our auto zoom algorithm first identifies and localizes the target, then crops the target region and scales it to the input size of the model, making it easier to extract the key features and fit the hardware resources. As target objects appear significantly smaller in aerial data due to high altitude, our auto zoom can automatically apply a zoom-in strategy, obtaining more action information from the target object. For example, a human actor in the UAV-Human dataset\cite{li2021uav} occupies 2\% to 5\% pixels of the frame. With our auto zoom algorithm, the pixels become about 16\% to 22\% of each frame. This provides more local details, making the model mainly focus on the actions rather than the background.
Moreover, our auto zoom can reduce the noise and outliers caused by the movement of a UAV. The zoom-in view changes every frame according to the position of the target object, which compromises the UAV's motion, so that the target object always appear in the center of the video. This extracts more useful features and makes the model more robust.


Current SOTA detectors use wide and deep network structures to obtain high performance, which require large computational resources while inferencing. To improve the efficiency while obtaining the same accuracy, our auto zoom algorithm goes through three steps:

    \noindent (1) The size of the cropping region changes in a dynamic manner. The height and width of the cropping region is chosen dynamically from $[480, 640, 720, 960]$. It depends not only on the raw video resolution, but also the size of the bboxes. We first calculate how many pixels belong to the bbox of the human actor, and then choose the size such that the human actor occupies 15\% to 20\% of the region. From our experiments, we find such a ratio is appropriate since it shows more details about the human actor as well as the surroundings. This ensures the model has enough information to deduce the relation between the human actor and the surroundings. Finally, we scale the target region to model input size for training.

    \noindent (2) We only use the detector for inferencing on key frames. Since most videos in UAV datasets are captured at high frame rates, only an extremely small portion of pixels' values will change in adjacent frames. It is unnecessary to obtain bboxes for each frame, as the results may only differ by a few pixels. Therefore, we only generate bboxes for 10\% or 20\% of the frames. Those frames are key frames and will be used as anchors for the bbox shifting in step 3. This can significantly reduce the computational cost and leave more resources for recognition.
    
    \noindent (3) For initialization, we uniformly distribute the key frames, giving a fixed stride. However, since the pretrained detector (e.g., Mobilenet, Cascade Mask RCNN) may not perform very well on the initial key frames, incorrect bboxes may be generated. To avoid this situation, we only take the bboxes which scores greater than 0.8 to filter out those with low possibility containing a human actor. Then we use Eq(\ref{eq:predict}) to predict the location of the bbox in the key frame~$t$ based on the bboxes generated from previous 3 key frame. 

    \begin{align}
    \begin{split}
    \label{eq:predict}
    &\begin{pmatrix}
            x_{t+1}\\
            y_{t+1}
        \end{pmatrix}
        =
        \begin{pmatrix}
            x_{t}+ (D_t + \delta_{D_t}) \cdot \cos{(\theta_{t-1}+\delta_{\theta})}\\
            y_{t}+ (D_t + \delta_{D_t}) \cdot \sin{(\theta_{t-1}+\delta_{\theta})}
        \end{pmatrix}, \\
    &\delta_{D_t} = D_{t} - D_{t-1} ,D_{t} = distance \left(\begin{pmatrix}
        x_{t-1}\\
        y_{t-1}
    \end{pmatrix},\begin{pmatrix}
        x_{t}\\
        y_{t}
    \end{pmatrix}\right), \\
    &\theta_{t-1} =  \arctan(\frac{y_{t}-y_{t-1}}{x_{t}-x_{t-1}}), \delta_{\theta} = \theta_{t-1} - \theta_{t-2}. 
    \end{split}
    \end{align}
  $x_t$ and $y_t$ are the coordinates of the bbox at key frame $t$. $D_{t}$, $\theta_{t}$ represent the shifting distance and angle between key frames $t-1$ and $t$, respectively. $\delta_{D_t}$ and $\delta_{theta}$ stand for deviations of the shfiting distance and angle at frame $t$.
  
  Next, we calculate the Euclidean distance between the positions of the predicted bbox and the bbox generated by the detector. If the distance is within the threshold we set, then we consider it the right detection result. If the distance is larger than the threshold or there is no detection result, we will assign the predicted bbox to the current key frame. For normal frames between key frames, we apply linear interpolation to get the bboxes.



\subsection{Temporal Reasoning}
We present the temporal reasoning algorithm to infer the action by analyzing the time sequence information. The temporal reasoning algorithm has three different kinds of implementation manner using 2D+1 convolution (conv), 3D conv and attention mechanism. The 2D+1 conv and attention mechanism can be used for the efficient model on edge devices. 3D conv and attention mechanism can be used for high performance model on decent GPUs. For 2D+1 conv temporal reasoning algorithm, we first utilize the 2D conv to generate features from the spatial information for each frame, and then perform 1D convolution to fuse all the spatial features so that the model can learn from the temporal information. As for 3D conv based temporal reasoning algorithm, 3D conv is used to handle spatial and temporal information simultaneously. Next, we present more details about the attention-based temporal reasoning algorithm, which includes cross-attention and self-attention.

Attention-based temporal reasoning algorithm consists of QKV(queries, keys and values) attention layers. The QKV attention mechanism encodes each input as a function of all the other inputs~\cite{vaswani2017attention,jaegle2021perceiver}. The output of attention function is computed as a weighted sum of the values, where the weight assigned to each value is computed by a compatibility function of the query with the corresponding key\cite{vaswani2017attention}. As shown in Fig. \ref{fig: temporal_reasoning}, attention-based temporal reasoning algorithm has two components, cross attention and self attention. The cross attention is used to map the input sequences to new sequence with a specific size according the computational cost requirement. The self attention is the standard component of transformers\cite{vaswani2017attention}. The difference of cross attention and self attention is cross attention has two inputs $X_Q \in \mathbb{R}^{N \times M}$ and $X_{KV} \in \mathbb{R}^{T \times D}$, and self attention has one input $X_{QKV} \in \mathbb{R}^{N \times M}$. $T, D, N, M$ donate input video frames, each frame's embedding dimension, query sequence number, each query sequence's dimension, respectively. For cross attention, $X_Q$ is the query input and $X_{KV}$ is the key and value input. For self attention, query, key and value share the same input $X_{QKV}$.  The input will though their projection function $p(\cdot)$ first,
\begin{equation}
F_Q, F_K, F_V = p(X_Q), p(X_{KV}), p(X_{KV}),
\end{equation}
where $F_Q \in \mathbb{R}^{T \times M}, F_K \in \mathbb{R}^{T \times S}, F_V \in \mathbb{R}^{N \times S}$ for cross attention and $F_Q \in \mathbb{R}^{N \times M}, F_K \in \mathbb{R}^{N \times S}, F_V \in \mathbb{R}^{N \times S}$ for self attention, where $F_Q$,$F_K$ and $F_V$ stand for the projected feature for the query, key and value. The projection function will map the key and value's dimension to $S$. Then we can adjust $S$ according to the information requirement and the computation limitation. Next, we calculate the attention,
\begin{equation}
\operatorname{X_{QK}=Attention}(Q, K)=\operatorname{softmax}\left(\frac{F_Q {F_K}^T}{\sqrt{S}}\right),
\end{equation}
Then we apply to the value,
\begin{align}
\begin{split}
\operatorname{Attention}\left(X_Q, X_{K V}\right)&=\operatorname{Attention}(Q, K, V) \\
&=X_{QK} V .
\end{split}
\end{align}
The input of the temporal reasoning algorithm comes from the auto zoom, and each sequence represent a frame's key information. Compared with CNN reasoning method, the attention based method can access the whole target information in the first layer of the model, which will offer more information in the whole process. And for the computational complexity, the dominated computation lies in the matrix multiply. For cross attention, we do multiply for matrix in shape of $T\times M, N\times T$ and $ T\times S, N\times S$, so the complexity can be expressed as $\mathcal{O}(TMNS)$. And for self attention, the complexity is $\mathcal{O}(MNS)$. If we have $L$ self-attention layer, the total complexity is $\mathcal{O}((T+L)MNS)$. Therefore, temporal reasoning algorithm has a linear complexity in terms of input temporal dimension $T$ and the model depth $L$. Besides, we also can control the computational cost by control the projection dimension $S$ for key and value.

\section{Experiments}
\label{sec: experiment}


In this section, we evaluate our algorithms and compare the performance with other state-of-the-art video action recognition methods on 3 UAV datasets. We evaluate the performance on edge devices as well as desktop GPUs.

\noindent  \textbf{Edge devices:} We use a robotic platform (Qualcomm Robotics RB5) with Qualcomm Kryo 585 CPU and Qualcomm Adreno 650. The efficient models are trained using TensorFlow and deployed using Robot Operating System 2 (ROS2) Galactic.

\noindent  \textbf{Desktop GPUs:} We use a high-end desktop with Intel Xeon W-2288 CPU and 8x Nvidia RTX A5000 GPUs. We train and test the high performance models using PyTorch.

\begin{table}
\centering
\resizebox{1.0\columnwidth}{!}{
\begin{tabular}{c c c c c}
\toprule
Platforms & C3D & MP3D &AP3D & DC3D   \\
\midrule
TL (CPU)\cite{tensorflow2015-whitepaper} &  \checkmark & $\times$ & $\times$ & $\times$ \\
TL + NNAPI (CPU/GPU/DSP)\cite{tensorflow2015-whitepaper} &  $\times$ & $\times$ & $\times$ & $\times$ \\
Tensorflow + MNN (CPU/GPU)\cite{tensorflow2015-whitepaper}\cite{proc:osdi22:walle} &   \checkmark &  \checkmark & \checkmark & $\times$ \\
TL + MNN\cite{tensorflow2015-whitepaper}\cite{proc:osdi22:walle} &  $\times$ & $\times$ & $\times$ & $\times$ \\
\bottomrule
\end{tabular}
}
\caption{3D operators are not well supported on  most edge devices or processors, as highlighted here. Therefore, we use 2D+1 conv and an efficient attention mechanism on the RB5 platform. TL: Tensorflow Lite, C3D: Conv3D, MP3D: MaxPooling 3D, AP3D: AveragePooling 3D, DC3D: Depthwidth Conv3D} 
\label{tab:platform}
\end{table}

\begin{table}
\centering
\resizebox{0.95\columnwidth}{!}{
\begin{tabular}{c c c c c c}
\toprule
Method & Input Size & Inference Time per frame  \\
\midrule 
MoViNet A0 \cite{kondratyuk2021movinets} & $172\times172$ & $33.2$ ms\\
MoViNet A2 \cite{kondratyuk2021movinets} & $224\times224$ & $106.4$ ms\\
MoViNet A3 \cite{kondratyuk2021movinets} & $256\times256$ &  $124.0$ ms \\

\textbf{AZTR (Ours)} & $172\times172$  & \textbf{56.5 ms} \\

\bottomrule 
\end{tabular}
}
\caption{Inference Time on RB5 CPU. Our method takes 56.5 ms to inference one frame (on average) which is 2 times faster than MoViNet A3 on the RB5, and also results in improvement on top-1 accuracy, see Table~\ref{tab:arl}. }
\label{tab:inf}
\end{table}
\begin{table}
\centering
\resizebox{0.95\columnwidth}{!}{
\begin{tabular}{c c c c c c}
\toprule
Method & Frames & Input Size & Init. & Top-1   \\
\midrule 
MoViNet A0 \cite{kondratyuk2021movinets} & $8$ & $172\times172$ & None & $17.8$ \\
MoViNet A0 \cite{kondratyuk2021movinets} & $8$ & $172\times172$ & Kinectics & $23.4$ \\
MoViNet A2 \cite{kondratyuk2021movinets} & $8$ & $224\times224$ & Kinectics & $28.1$ \\
MoViNet A3 \cite{kondratyuk2021movinets} & $8$ & $256\times256$ & Kinectics & $29.0$ \\
\textbf{AZTR (Ours)} & $8$ & $172\times 172$ & Kinectics & \textbf{29.5} \\
\midrule 
MoViNet A0 \cite{kondratyuk2021movinets} & $20$ & $172\times172$ & None & $27.5$ \\
MoViNet A0 \cite{kondratyuk2021movinets} & $20$ & $172\times172$ & Kinectics & $32.8$ \\
MoViNet A2 \cite{kondratyuk2021movinets} & $20$ & $224\times224$ & Kinectics & $34.1$ \\
\textbf{AZTR (Ours)} & $20$ & $172\times 172$ & Kinectics & \textbf{40.2} \\

\bottomrule 
\end{tabular}
}
\caption{Results on RoCoG-v2. We demonstrate that our approach can improve the top-1 accuracy by 6.1\%-7.4\%, outperforms all SOTA methods that can be deployed on the RB5 platform.}
\vspace{-15pt}
\label{tab:arl}
\end{table}

\begin{figure}[t]
    \centering
        \includegraphics[width=\columnwidth]{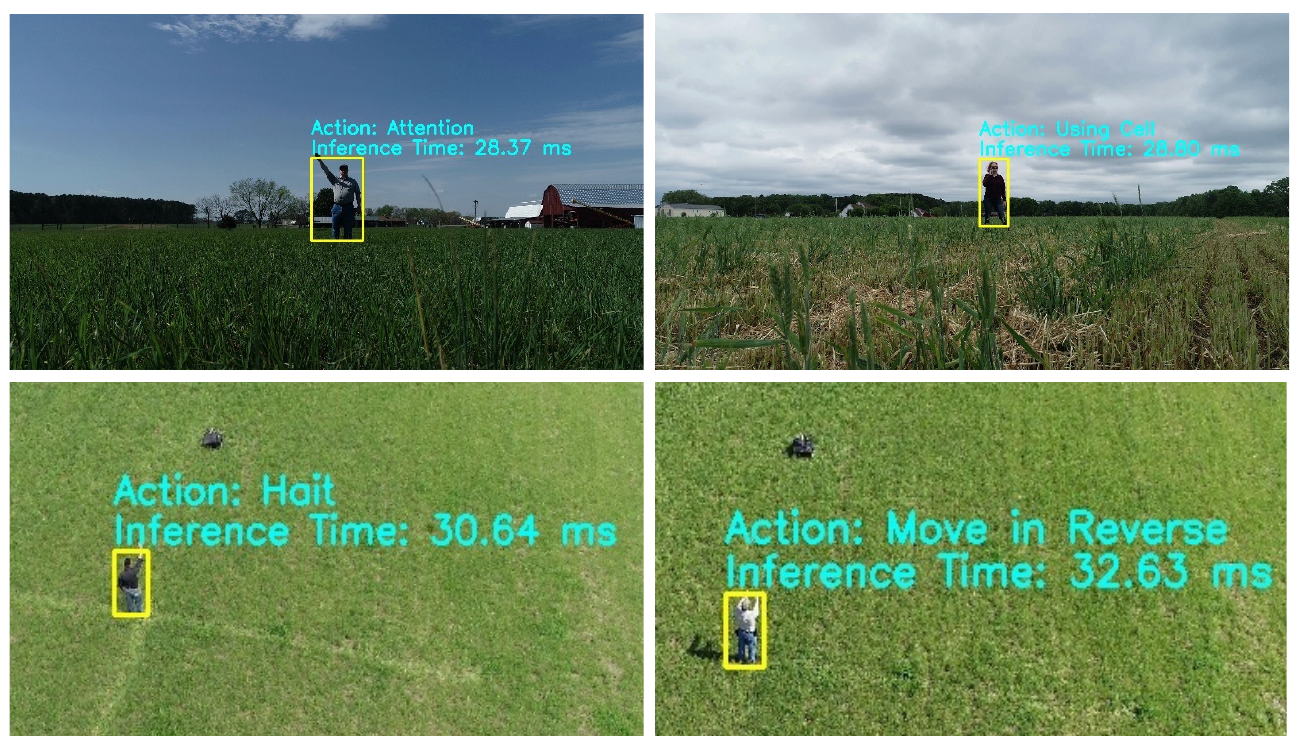}
        \caption{Action recognition on RoCoG-v2 aerial video.  More details are given in the video.}
        \vspace{-10pt}
    \label{fig: visual}
\end{figure}

\subsection{Datasets}

\noindent  \textbf{RoCoG-v2~\cite{de2022icras,de2020vision}:} We use 99 long raw videos with 17 action categories. We trim the original videos into 5,828 1-5 seconds videos clips, with one action category per clip. We also obtained 70,124 frames with bounding boxes in a semi-automatic manner. We first use a human detector (Cascaded Mask RCNN) to compute the coarse bounding box, followed by human manual check. We also fine-tune our detection model on RoCoG-v2, which can narrow the domain gap for the detection. Overall, we achieve 6.1-7.4\% accuracy improvement using our method on this dataset.

\noindent  \textbf{UAV Human~\cite{li2021uav}:} It is one of the largest UAV-based human behavior understanding dataset. It contains 15172 and 5556 videos for training and testing, respectively. All the videos are collected in multiple urban and rural areas in both day and night  settings with extensive diversity (w.r.t subjects, backgrounds, illuminations, etc.). Our method improves the accuracy by 8.3-10.4\% on the UAV-Human over SOTA.

\noindent  \textbf{Drone Action~\cite{perera2019drone}:} An outdoor drone video dataset captured using a free-flying drone at low altitudes and low speeds. It has 240 HD RGB videos across 13 human actions. Our approach outperforms SOTA by 3.2\%, reaches 95.9\% Top-1 accuracy on the Drone Action.

\subsection{Implementation Details and Training}
\noindent \textbf{Backbone network architectures:} We benchmark our models by using Movinets\cite{kondratyuk2021movinets} as the backbone for lightweight model and using X3D-M \cite{feichtenhofer2020x3d} as the backbone for models on desktop GPUs.

\noindent \textbf{Localization:} We applied Mobilenet v2\cite{sandler2018mobilenetv2} as the locator for lightweight models, and use Cascaded Mask RCNN\cite{hasan2021generalizable} as the locator for models on desktop GPUs.

\noindent \textbf{Training details:} For models on edge devices or UAV platforms, we set $0.00005$ as the initial learning rate. For models on high-end desktop GPUs, we use the same initialization as in~\cite{kothandaraman2022fourier}. We set 0.1 as the initial learning rate for training from scratch and 0.05 for initializing with Kinetics pretrained weights. We use the Stochastic Gradient Descent (SGD) optimizer with weight decay of $0.0005$ and momentum of $0.9$ and cosine/poly annealing for learning rate decay. All our models were trained using NVIDIA GeForce 2080 Ti GPUs and NVIDIA RTX A5000 GPUs.


\subsection{Results on RoCoG-v2}

We implement our methods on the RoCoG-v2 dataset in a preinstalled Ubuntu Linux 18.04 using Robot Operating System 2(ROS2) Galactics. We fine-tune a MobileNet v2\cite{sandler2018mobilenetv2} using the annotated 70,124 frames extracted from raw video data. The fine-tuned MobileNet v2\cite{sandler2018mobilenetv2} is operated as the core detector in our auto zoom algorithm. We evaluate our method with Mobile Video Networks(MoViNets)\cite{kondratyuk2021movinets}. Specifically, we test our method with MoViNet A0, A2 and A3 on RB5. MoViNet A0 is the smallest model in MoViNets and has input size 172$\times$172. A2 uses deeper network design with input 224$\times$224. A3 is the larger network, with input 256$\times$256. Also, we choose the stream version of the MoViNets due to lack of support for 3D conv on edge devices, as shown in Table~\ref{tab:platform}. The 2D + 1D convolution and a stream buffer is used to fuse the recognition results from previous frames, making it capable of online inferencing. We deploy MoViNet A0, A2, A3 and our method on the Qualcomm RB5 platform. We collect the onboard inference time on RB5 CPU. The result is shown in Table~\ref{tab:inf}. Our method obtains better speed-accuracy trade-off on mobile devices. As compared with MoViNet A3, the average inference time for our method on RB5 CPU is 56.5 ms, which is 2X faster, and results in slight accuracy improvement. We performed more experiments with different frame sampling and model initialization to evaluate the performance of our algorithm (see Table~\ref{tab:arl}). With the same configuration, our method achieves 6.1-7.4\% improvement on the RoCoG-v2 dataset.

\subsection{Results on UAV Human}

\begin{table}
\resizebox{1.0\columnwidth}{!}{
\begin{tabular}{c c c c c c}
\toprule
Method & Backbone & Frames & Input Size & Init. & Top-1   \\
\midrule 
X3D-S \cite{feichtenhofer2020x3d}& - & $16$ & $224\times224$ & None & $21.5$ \\
X3D-M \cite{feichtenhofer2020x3d}& - & $16$ & $224\times224$ & None & $27.0$ \\
X3D-L \cite{feichtenhofer2020x3d}& - & $16$ & $224\times224$ & None & $27.6$ \\
FAR \cite{kothandaraman2022fourier} & X3D-M & $16$ & $224\times 224$ & None & $27.6$ \\
FAR \cite{kothandaraman2022fourier} & X3D-M & $8$ & $540\times 540$ & None & $28.8$ \\
\textbf{AZTR (Ours)}& X3D-M & $16$ & $224\times 224$ & None & \textbf{39.2} \\
\midrule 
I3D \cite{carreira2017quo}& ResNet-101 & $8$ & $540\times960$ & Kinetics & $21.06$ \\
X3D-M \cite{feichtenhofer2020x3d}& - & $16$ & $224\times224$ & Kinetics & $30.6$ \\
FNet \cite{lee2021fnet} & I3D & $8$ & $540\times960$ & Kinetics & $24.39$ \\
FAR \cite{kothandaraman2022fourier} & I3D & $8$ & $540\times960$ & Kinetics & $29.21$ \\
FAR \cite{kothandaraman2022fourier} & X3D-M & $16$ & $224 \times 224$ & Kinetics & $31.9$ \\
FAR \cite{kothandaraman2022fourier} & X3D-M & $8$ & $620\times620$ & Kinetics & $39.1$ \\
\textbf{AZTR (Ours)} & X3D-M & $16$ & $224\times 224$ & Kinetics & \textbf{47.4} \\
\bottomrule 
\end{tabular}
}
\caption{Benchmarking UAV Human and comparisons with prior arts. We compared with the state-of-the-art methods, which demonstrates an improvement of $8.3\%-10.4$ over SOTA methods. Trained on high-end desktop GPUs.}
\vspace{-5pt}
\label{tab:uavhuman_sota}
\end{table}
To illustrate the effectiveness of our proposed method, we evaluate our algorithms on the largest UAV dataset UAV-Human~\cite{li2021uav}. As shown in Table~\ref{tab:uavhuman_sota}, we conduct experiments on UAV-Human in terms of different backbone network architectures, frame rates, input sizes, and weights initialization. We use X3D-M as backbone with two different initialization settings in our experiments, one is to train from scratch and the other is to initial the weights with Kinetics pretrained weights. Based on our results, we observe that initializing with Kinetics pre-trained weights outperforms training from scratch. Furthermore, compared with the SOTA methods' large input size, we achieved considerable improvement even with smaller input size( $224\times 224$). For training from scratch, our AZTR improves performance over the baseline by 11.6\% and over the state-of-the-art by 10.4\%. For initializing with Kinetics pretrained weights, AZTR improves performance over the baseline by 16.8\% and over the state-of-the-art by 8.3\%. 

\subsection{Results on Drone Action}
\begin{table}
\centering
\resizebox{1.0\columnwidth}{!}{
\begin{tabular}{c c c c c}
\toprule
Method & Frames & Input Size & Init. & Top-1   \\
\midrule
HLPF \cite{jhuang2013towards}&  All & $1920\times1080$ & None & $64.36$ \\
PCNN \cite{cheron2015p} &  - & $1920\times1080$ & None & $75.92$\\
X3D-M \cite{feichtenhofer2020x3d}&  $16$ & $224\times224$ & Kinetics & $83.4$ \\
FAR \cite{kothandaraman2022fourier} &  $16$ & $224\times224$ & Kinetics & $92.7$ \\
\textbf{AZTR (Ours)} &  $16$ & $224\times224$ & Kinetics & \textbf{95.9} \\
\bottomrule
\end{tabular}
}
\caption{Results on dataset: Drone Action. We demonstrate that AZTR improves the state-of-the-art accuracy by 3.2\%, reaching 95.9\% on Drone Action. Trained on high-end desktop GPUs.}
\vspace{-10pt}
\label{tab:more_drones}
\end{table}

We also evaluated our method on Drone Action~\cite{perera2019drone}. As shown in Table~\ref{tab:more_drones}, we demonstrate that AZTR outperforms SOTA by $3.2\%$ on Drone Action with 95.9\% Top-1 accuracy, which illustrate the stable performance and the convincingness of our proposed AZTR.

\section{Conclusion, Limitations and Future}
\label{sec: conclusion}

We present a novel approach for video action recognition. Our approach is designed for edge devices used in UAVs (e.g., Qualcomm Robotics RB5) and desktop GPUs (Nvidia RTX A5000 GPUs). 
We present auto zoom algorithm that can automatically identify and scale the target. Additionally, we introduce a temporal reasoning algorithm to capture the action information. We observe improvements in accuracy on different datasets. Our method has a few limitations. First, the overall performance depends on the localization methods. Second, we assume our input videos only has one scripted human actor performing. We would also like to develop methods that can perform action recognition in videos with multiple human actors and can handling varying lighting and weather conditions.\\

\bibliographystyle{IEEEtran}
\bibliography{refs}

\begin{thebibliography}{10}
\providecommand{\url}[1]{#1}
\csname url@samestyle\endcsname
\providecommand{\newblock}{\relax}
\providecommand{\bibinfo}[2]{#2}
\providecommand{\BIBentrySTDinterwordspacing}{\spaceskip=0pt\relax}
\providecommand{\BIBentryALTinterwordstretchfactor}{4}
\providecommand{\BIBentryALTinterwordspacing}{\spaceskip=\fontdimen2\font plus
\BIBentryALTinterwordstretchfactor\fontdimen3\font minus
  \fontdimen4\font\relax}
\providecommand{\BIBforeignlanguage}[2]{{%
\expandafter\ifx\csname l@#1\endcsname\relax
\typeout{** WARNING: IEEEtran.bst: No hyphenation pattern has been}%
\typeout{** loaded for the language `#1'. Using the pattern for}%
\typeout{** the default language instead.}%
\else
\language=\csname l@#1\endcsname
\fi
#2}}
\providecommand{\BIBdecl}{\relax}
\BIBdecl

\bibitem{nguyen2022state}
K.~Nguyen, C.~Fookes, S.~Sridharan, Y.~Tian, X.~Liu, F.~Liu, and A.~Ross, ``The
  state of aerial surveillance: A survey,'' \emph{arXiv preprint
  arXiv:2201.03080}, 2022.

\bibitem{pan2022edgevits}
J.~Pan, A.~Bulat, F.~Tan, X.~Zhu, L.~Dudziak, H.~Li, G.~Tzimiropoulos, and
  B.~Martinez, ``Edgevits: Competing light-weight cnns on mobile devices with
  vision transformers,'' \emph{arXiv preprint arXiv:2205.03436}, 2022.

\bibitem{kondratyuk2021movinets}
D.~Kondratyuk, L.~Yuan, Y.~Li, L.~Zhang, M.~Tan, M.~Brown, and B.~Gong,
  ``Movinets: Mobile video networks for efficient video recognition,'' in
  \emph{IEEE/CVF Conference on Computer Vision and Pattern Recognition (CVPR)},
  2021, pp. 16\,020--16\,030.

\bibitem{demir2021tinyvirat}
U.~Demir, Y.~S. Rawat, and M.~Shah, ``Tinyvirat: Low-resolution video action
  recognition,'' in \emph{2020 25th International Conference on Pattern
  Recognition (ICPR)}.\hskip 1em plus 0.5em minus 0.4em\relax IEEE, 2021, pp.
  7387--7394.

\bibitem{rezazadegan2017action}
F.~Rezazadegan, S.~Shirazi, B.~Upcrofit, and M.~Milford, ``Action recognition:
  From static datasets to moving robots,'' in \emph{2017 IEEE International
  Conference on Robotics and Automation (ICRA)}.\hskip 1em plus 0.5em minus
  0.4em\relax IEEE, 2017, pp. 3185--3191.

\bibitem{soans2020sa}
N.~Soans, E.~Asali, Y.~Hong, and P.~Doshi, ``Sa-net: Robust state-action
  recognition for learning from observations,'' in \emph{2020 IEEE
  International Conference on Robotics and Automation (ICRA)}.\hskip 1em plus
  0.5em minus 0.4em\relax IEEE, 2020, pp. 2153--2159.

\bibitem{van2020multi}
B.~van Amsterdam, M.~J. Clarkson, and D.~Stoyanov, ``Multi-task recurrent
  neural network for surgical gesture recognition and progress prediction,'' in
  \emph{2020 IEEE International Conference on Robotics and Automation
  (ICRA)}.\hskip 1em plus 0.5em minus 0.4em\relax IEEE, 2020, pp. 1380--1386.

\bibitem{massardi2020parc}
J.~Massardi, M.~Gravel, and {\'E}.~Beaudry, ``Parc: A plan and activity
  recognition component for assistive robots,'' in \emph{2020 IEEE
  International Conference on Robotics and Automation (ICRA)}.\hskip 1em plus
  0.5em minus 0.4em\relax IEEE, 2020, pp. 3025--3031.

\bibitem{yao2022pa}
L.~Yao, S.~Liu, C.~Li, S.~Zou, S.~Chen, and D.~Guan, ``Pa-awcnn: Two-stream
  parallel attention adaptive weight network for rgb-d action recognition,'' in
  \emph{2022 International Conference on Robotics and Automation (ICRA)}.\hskip
  1em plus 0.5em minus 0.4em\relax IEEE, 2022, pp. 8741--8747.

\bibitem{shao2018hierarchical}
Z.~Shao, Y.~Li, Y.~Guo, J.~Yang, and Z.~Wang, ``A hierarchical model for action
  recognition based on body parts,'' in \emph{2018 IEEE international
  conference on robotics and automation (ICRA)}.\hskip 1em plus 0.5em minus
  0.4em\relax IEEE, 2018, pp. 1978--1985.

\bibitem{lea2016learning}
C.~Lea, R.~Vidal, and G.~D. Hager, ``Learning convolutional action primitives
  for fine-grained action recognition,'' in \emph{2016 IEEE international
  conference on robotics and automation (ICRA)}.\hskip 1em plus 0.5em minus
  0.4em\relax IEEE, 2016, pp. 1642--1649.

\bibitem{geraldes2019uav}
R.~Geraldes, A.~Goncalves, T.~Lai, M.~Villerabel, W.~Deng, A.~Salta,
  K.~Nakayama, Y.~Matsuo, and H.~Prendinger, ``Uav-based situational awareness
  system using deep learning,'' \emph{IEEE Access}, vol.~7, pp.
  122\,583--122\,594, 2019.

\bibitem{mliki2020human}
H.~Mliki, F.~Bouhlel, and M.~Hammami, ``Human activity recognition from
  uav-captured video sequences,'' \emph{Pattern Recognition (PR)}, vol. 100, p.
  107140, 2020.

\bibitem{mishra2020drone}
B.~Mishra, D.~Garg, P.~Narang, and V.~Mishra, ``Drone-surveillance for search
  and rescue in natural disaster,'' \emph{Computer Communications}, vol. 156,
  pp. 1--10, 2020.

\bibitem{mou2020event}
L.~Mou, Y.~Hua, P.~Jin, and X.~X. Zhu, ``Event and activity recognition in
  aerial videos using deep neural networks and a new dataset,'' in \emph{IGARSS
  2020-2020 IEEE International Geoscience and Remote Sensing Symposium
  (IGARSS)}.\hskip 1em plus 0.5em minus 0.4em\relax IEEE, 2020, pp. 952--955.

\bibitem{barekatain2017okutama}
M.~Barekatain, M.~Mart{\'\i}, H.-F. Shih, S.~Murray, K.~Nakayama, Y.~Matsuo,
  and H.~Prendinger, ``Okutama-action: An aerial view video dataset for
  concurrent human action detection,'' in \emph{IEEE/CVF Conference on Computer
  Vision and Pattern Recognition Workshops (CVPRW)}, 2017, pp. 28--35.

\bibitem{perera2019drone}
A.~G. Perera, Y.~W. Law, and J.~Chahl, ``Drone-action: An outdoor recorded
  drone video dataset for action recognition,'' \emph{Drones}, vol.~3, no.~4,
  p.~82, 2019.

\bibitem{perera2020multiviewpoint}
A.~G. Perera, Y.~W. Law, T.~T. Ogunwa, and J.~Chahl, ``A multiviewpoint outdoor
  dataset for human action recognition,'' \emph{IEEE Transactions on
  Human-Machine Systems}, vol.~50, no.~5, pp. 405--413, 2020.

\bibitem{choi2020unsupervised}
J.~Choi, G.~Sharma, M.~Chandraker, and J.-B. Huang, ``Unsupervised and
  semi-supervised domain adaptation for action recognition from drones,'' in
  \emph{IEEE/CVF Winter Conference on Applications of Computer Vision (WACV)},
  2020, pp. 1717--1726.

\bibitem{li2021uav}
T.~Li, J.~Liu, W.~Zhang, Y.~Ni, W.~Wang, and Z.~Li, ``Uav-human: A large
  benchmark for human behavior understanding with unmanned aerial vehicles,''
  in \emph{IEEE/CVF Conference on Computer Vision and Pattern Recognition
  (CVPR)}, 2021, pp. 16\,266--16\,275.

\bibitem{sultani2021human}
W.~Sultani and M.~Shah, ``Human action recognition in drone videos using a few
  aerial training examples,'' \emph{Computer Vision and Image Understanding},
  vol. 206, p. 103186, 2021.

\bibitem{carreira2017quo}
J.~Carreira and A.~Zisserman, ``Quo vadis, action recognition? a new model and
  the kinetics dataset,'' in \emph{IEEE/CVF Conference on Computer Vision and
  Pattern Recognition (CVPR)}, 2017, pp. 6299--6308.

\bibitem{peng2020fully}
H.~Peng and A.~Razi, ``Fully autonomous uav-based action recognition system
  using aerial imagery,'' in \emph{International Symposium on Visual Computing
  (ISVC)}.\hskip 1em plus 0.5em minus 0.4em\relax Springer, 2020, pp. 276--290.

\bibitem{kothandaraman2022fourier}
D.~Kothandaraman, T.~Guan, X.~Wang, S.~Hu, M.~Lin, and D.~Manocha, ``Far:
  Fourier aerial video recognition,'' \emph{arXiv preprint arXiv:2203.10694},
  2022.

\bibitem{du2018unmanned}
D.~Du, Y.~Qi, H.~Yu, Y.~Yang, K.~Duan, G.~Li, W.~Zhang, Q.~Huang, and Q.~Tian,
  ``The unmanned aerial vehicle benchmark: Object detection and tracking,'' in
  \emph{European Conference on Computer Vision (ECCV)}, 2018, pp. 370--386.

\bibitem{zhu2020detection}
P.~Zhu, L.~Wen, D.~Du, X.~Bian, H.~Fan, Q.~Hu, and H.~Ling, ``Detection and
  tracking meet drones challenge,'' \emph{arXiv preprint arXiv:2001.06303},
  2020.

\bibitem{perera2018uav}
A.~G. Perera, Y.~Wei~Law, and J.~Chahl, ``Uav-gesture: A dataset for uav
  control and gesture recognition,'' in \emph{European Conference on Computer
  Vision Workshops (ECCVW)}, 2018, pp. 0--0.

\bibitem{de2020vision}
C.~M. de~Melo, B.~Rothrock, P.~Gurram, O.~Ulutan, and B.~Manjunath,
  ``Vision-based gesture recognition in human-robot teams using synthetic
  data,'' in \emph{2020 IEEE/RSJ International Conference on Intelligent Robots
  and Systems (IROS)}.\hskip 1em plus 0.5em minus 0.4em\relax IEEE, pp.
  10\,278--10\,284.

\bibitem{fan2019more}
Q.~Fan, C.-F.~R. Chen, H.~Kuehne, M.~Pistoia, and D.~Cox, ``More is less:
  Learning efficient video representations by big-little network and depthwise
  temporal aggregation,'' \emph{Advances in Neural Information Processing
  Systems (NeurIPS)}, vol.~32, 2019.

\bibitem{bhardwaj2019efficient}
S.~Bhardwaj, M.~Srinivasan, and M.~M. Khapra, ``Efficient video classification
  using fewer frames,'' in \emph{IEEE/CVF Conference on Computer Vision and
  Pattern Recognition (CVPR)}, 2019, pp. 354--363.

\bibitem{chen2018big}
C.-F. Chen, Q.~Fan, N.~Mallinar, T.~Sercu, and R.~Feris, ``Big-little net: An
  efficient multi-scale feature representation for visual and speech
  recognition,'' \emph{arXiv preprint arXiv:1807.03848}, 2018.

\bibitem{li2020smallbignet}
X.~Li, Y.~Wang, Z.~Zhou, and Y.~Qiao, ``Smallbignet: Integrating core and
  contextual views for video classification,'' in \emph{IEEE/CVF Conference on
  Computer Vision and Pattern Recognition (CVPR)}, 2020, pp. 1092--1101.

\bibitem{tran2019video}
D.~Tran, H.~Wang, L.~Torresani, and M.~Feiszli, ``Video classification with
  channel-separated convolutional networks,'' in \emph{IEEE/CVF International
  Conference on Computer Vision (ICCV)}, 2019, pp. 5552--5561.

\bibitem{howard2017mobilenets}
A.~G. Howard, M.~Zhu, B.~Chen, D.~Kalenichenko, W.~Wang, T.~Weyand,
  M.~Andreetto, and H.~Adam, ``Mobilenets: Efficient convolutional neural
  networks for mobile vision applications,'' \emph{arXiv preprint
  arXiv:1704.04861}, 2017.

\bibitem{redmon2016you}
J.~Redmon, S.~Divvala, R.~Girshick, and A.~Farhadi, ``You only look once:
  Unified, real-time object detection,'' in \emph{IEEE/CVF Conference on
  Computer Vision and Pattern Recognition (CVPR)}, 2016, pp. 779--788.

\bibitem{ding2020lightweight}
M.~Ding, N.~Li, Z.~Song, R.~Zhang, X.~Zhang, and H.~Zhou, ``A lightweight
  action recognition method for unmanned-aerial-vehicle video,'' in \emph{2020
  IEEE 3rd International Conference on Electronics and Communication
  Engineering (ICECE)}.\hskip 1em plus 0.5em minus 0.4em\relax IEEE, 2020, pp.
  181--185.

\bibitem{piergiovanni2022tiny}
A.~Piergiovanni, A.~Angelova, and M.~S. Ryoo, ``Tiny video networks,''
  \emph{Applied AI Letters}, vol.~3, no.~1, p. e38, 2022.

\bibitem{lin2019tsm}
J.~Lin, C.~Gan, and S.~Han, ``Tsm: Temporal shift module for efficient video
  understanding,'' in \emph{IEEE/CVF International Conference on Computer
  Vision (ICCV)}, 2019, pp. 7083--7093.

\bibitem{vaswani2017attention}
A.~Vaswani, N.~Shazeer, N.~Parmar, J.~Uszkoreit, L.~Jones, A.~N. Gomez,
  {\L}.~Kaiser, and I.~Polosukhin, ``Attention is all you need,''
  \emph{Advances in Neural Information Processing Systems (NeurIPS)}, vol.~30,
  2017.

\bibitem{jaegle2021perceiver}
A.~Jaegle, F.~Gimeno, A.~Brock, O.~Vinyals, A.~Zisserman, and J.~Carreira,
  ``Perceiver: General perception with iterative attention,'' in
  \emph{International Conference on Machine Learning (ICML)}.\hskip 1em plus
  0.5em minus 0.4em\relax PMLR, 2021, pp. 4651--4664.

\bibitem{tensorflow2015-whitepaper}
\BIBentryALTinterwordspacing
M.~Abadi, A.~Agarwal, P.~Barham, E.~Brevdo, Z.~Chen, C.~Citro, G.~S. Corrado,
  A.~Davis, J.~Dean, M.~Devin, S.~Ghemawat, I.~Goodfellow, A.~Harp, G.~Irving,
  M.~Isard, Y.~Jia, R.~Jozefowicz, L.~Kaiser, M.~Kudlur, J.~Levenberg,
  D.~Man\'{e}, R.~Monga, S.~Moore, D.~Murray, C.~Olah, M.~Schuster, J.~Shlens,
  B.~Steiner, I.~Sutskever, K.~Talwar, P.~Tucker, V.~Vanhoucke, V.~Vasudevan,
  F.~Vi\'{e}gas, O.~Vinyals, P.~Warden, M.~Wattenberg, M.~Wicke, Y.~Yu, and
  X.~Zheng, ``{TensorFlow}: Large-scale machine learning on heterogeneous
  systems,'' 2015, software available from tensorflow.org. [Online]. Available:
  \url{https://www.tensorflow.org/}
\BIBentrySTDinterwordspacing

\bibitem{proc:osdi22:walle}
\BIBentryALTinterwordspacing
C.~Lv, C.~Niu, R.~Gu, X.~Jiang, Z.~Wang, B.~Liu, Z.~Wu, Q.~Yao, C.~Huang,
  P.~Huang, T.~Huang, H.~Shu, J.~Song, B.~Zou, P.~Lan, G.~Xu, F.~Wu, S.~Tang,
  F.~Wu, and G.~Chen, ``Walle: An {End-to-End}, {General-Purpose}, and
  {Large-Scale} production system for {Device-Cloud} collaborative machine
  learning,'' in \emph{16th USENIX Symposium on Operating Systems Design and
  Implementation (OSDI 22)}.\hskip 1em plus 0.5em minus 0.4em\relax Carlsbad,
  CA: USENIX Association, Jul. 2022, pp. 249--265. [Online]. Available:
  \url{https://www.usenix.org/conference/osdi22/presentation/lv}
\BIBentrySTDinterwordspacing

\bibitem{de2022icras}
A.~V. Reddy, K.~Shah, W.~Paul, R.~Mocharla, J.~Hoffman, K.~D. Katyal, v~Dinesh,
  C.~M. de~Melo, and R.~Chellappa, ``Rocog-v2,'' \emph{Under Review,
  https://github.com/reddyav1/RoCoG-v2}, 2022.

\bibitem{feichtenhofer2020x3d}
C.~Feichtenhofer, ``X3d: Expanding architectures for efficient video
  recognition,'' in \emph{IEEE/CVF Conference on Computer Vision and Pattern
  Recognition (CVPR)}, 2020, pp. 203--213.

\bibitem{sandler2018mobilenetv2}
M.~Sandler, A.~Howard, M.~Zhu, A.~Zhmoginov, and L.-C. Chen, ``Mobilenetv2:
  Inverted residuals and linear bottlenecks,'' in \emph{IEEE/CVF Conference on
  Computer Vision and Pattern Recognition (CVPR)}, 2018, pp. 4510--4520.

\bibitem{hasan2021generalizable}
I.~Hasan, S.~Liao, J.~Li, S.~U. Akram, and L.~Shao, ``Generalizable pedestrian
  detection: The elephant in the room,'' in \emph{IEEE/CVF Conference on
  Computer Vision and Pattern Recognition (CVPR)}, 2021, pp. 11\,328--11\,337.

\bibitem{lee2021fnet}
J.~Lee-Thorp, J.~Ainslie, I.~Eckstein, and S.~Ontanon, ``Fnet: Mixing tokens
  with fourier transforms,'' \emph{arXiv preprint arXiv:2105.03824}, 2021.

\bibitem{jhuang2013towards}
H.~Jhuang, J.~Gall, S.~Zuffi, C.~Schmid, and M.~J. Black, ``Towards
  understanding action recognition,'' in \emph{IEEE/CVF International
  Conference on Computer Vision (ICCV)}, 2013, pp. 3192--3199.

\bibitem{cheron2015p}
G.~Ch{\'e}ron, I.~Laptev, and C.~Schmid, ``P-cnn: Pose-based cnn features for
  action recognition,'' in \emph{IEEE/CVF International Conference on Computer
  Vision (ICCV)}, 2015, pp. 3218--3226.

\end{thebibliography}

\end{document}